\documentclass{article}
\usepackage{ijcai25}

\usepackage{times}
\usepackage{soul}
\usepackage{url}
\usepackage[hidelinks]{hyperref}
\usepackage[utf8]{inputenc}
\usepackage[small]{caption}
\usepackage{graphicx}
\usepackage{amsmath}
\usepackage{amsthm}
\usepackage{booktabs}
\usepackage{algorithm}
\usepackage{algorithmic}
\usepackage[switch]{lineno}
\usepackage{multirow}
\usepackage{xcolor,colortbl}

\title{RAVU: Retrieval Augmented Video Understanding \\with Compositional Reasoning over Graph}

\author{
Sameer Malik
\and
Moyuru Yamada\and
Ayush Singh\And
Dishank Aggarwal\\
\affiliations
Fujitsu Research of India Private Limited\\
\emails
\{sameer.malik, yamada.moyuru, ayush.singh, dishank.aggarwal\}@fujitsu.com
}

\begin{document}

\maketitle

\begin{abstract}
Comprehending long videos remains a significant challenge for Large Multi-modal Models (LMMs). Current LMMs struggle to process even minutes to hours videos due to their lack of explicit memory and retrieval mechanisms. To address this limitation, we propose \textbf{RAVU} (\textbf{R}etrieval \textbf{A}ugmented \textbf{V}ideo \textbf{U}nderstanding), a novel framework for video understanding enhanced by retrieval with compositional reasoning over a spatio-temporal graph. We construct a graph representation of the video, capturing both spatial and temporal relationships between entities. This graph serves as a long-term memory, allowing us to track objects and their actions across time. To answer complex queries, we decompose the queries into a sequence of reasoning steps and execute these steps on the graph, retrieving relevant key information. Our approach enables more accurate understanding of long videos, particularly for queries that require multi-hop reasoning and tracking objects across frames. Our approach demonstrate superior performances with limited retrieved frames (5-10) compared with other SOTA methods and baselines on two major video QA datasets, NExT-QA and EgoSchema.
\end{abstract}
\section{Introduction}
\label{sec:introduction}

Understanding videos inherently requires the ability to memorize multi-modal information and retrieve it according to a given task. Recent advancements in Large Multi-modal Models (LMMs) have shown promise in tackling this challenge ~\cite{moviechat,malmm,VideoAgent}. However, comprehending long videos, particularly multi-hop reasoning tasks, remains a significant challenge, even for these powerful models.

\begin{figure}[t]
    \centering
    \includegraphics[width=0.48\textwidth]{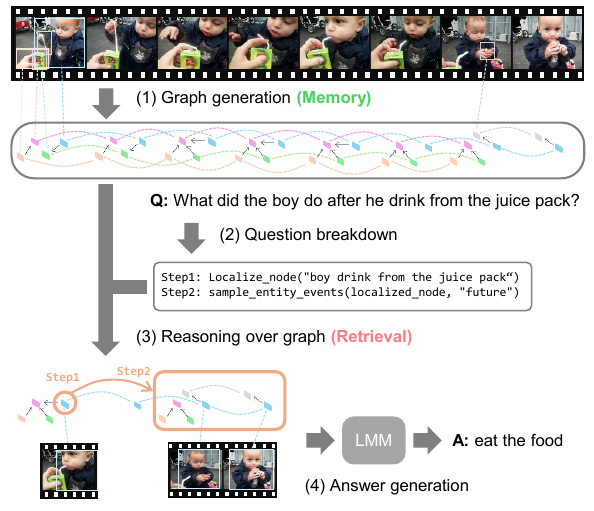 }
    \caption{Overview of RAVU, memorizing a video as a spatio-temporal graph and retrieving key relevant parts by reasoning over the graph with a given query.}
    \label{fig:overview}
\end{figure}

This limitation primarily stems from the absence of explicit memory and retrieval mechanisms in current Transformer-based LMMs. The current LMMs represent each video frame as hundreds of tokens and thus have difficulty in processing hours of video content. Even at 64 tokens per frame an hour-long video could require over 200k tokens ~\cite{longvu}. Without retrieval mechanism, they need to take the entire video as input even for questions about a specific part of a video. While some studies have explored constructing a long-term memory from an input video ~\cite{moviechat,malmm}, these approaches either sample key frames from the video or compress the video by grouping similar frames, regardless of the input query, potentially overlooking crucial details for the specific queries. Other methods ~\cite{moviechat_p} have proposed constructing long-term memories based on query relevance but require recompressing the video for each query. Additionally, agentic approaches ~\cite{VideoAgent} have been explored, where relevant frames are iteratively retrieved until sufficient information is obtained to answer the query. These existing approaches highly rely on simple similarity between the query and individual frames, lacking the capability to track the identity of objects across consecutive frames. For instance, they may fail to correctly identify which man in the previous frame corresponds to the man performing a specific action in the current frame. Such temporal connections are essential for accurately understanding videos with complex queries, such as "How does the girl in red react after being pulled backwards by the girl in blue?". These queries also necessitate multi-hop reasoning and may require more than simple frame-level relevance to identify important scenes in the long video.

The video is often represented as a graph in the fields of interaction detection ~\cite{video_sgg_1,video_sgg_2}, where nodes correspond to objects and edges represent the interaction between the objects. The graph then evolves through time. Some studies ~\cite{video-of-thought,graph_video} have initiated explorations of using graph for video understanding. However their models are trained on a specific datasets and may struggle for generalization.

To address these limitations, this paper proposes \textbf{RAVU} (Retrieval Augmented Video Understanding), a novel framework for video understanding based on compositional reasoning over a spatio-temporal graph. 
We first construct a spatio-temporal graph from the video with the LMM. This graph is generated once per the video, independent of the queries, and serves as a memory. In this memory, the same entities (e.g., a man and a dog) are connected across frames, allowing us to track the actions of specific individuals over time. Unlike the conventional methods which simply retrieve the video frames based on the similarity between the query and frames, we first decompose the complex query into reasoning steps and then retrieve the necessary scenes for answering the query by performing each reasoning step on the graph sequentially. While we implement various reasoning steps to cover a wide range of queries, it is possible to employ neural networks for each step. Furthermore, our memory and retrieval framework can be readily applied to existing LMMs without fine-tuning or can be fine-tuned on open-source LMMs.

The main contributions of our paper can be summarized as follows:

\begin{itemize}
    \item We propose a novel video understanding framework, which constructs a spatio-temporal graph as a memory from a video and retrieves key frames from the video by reasoning over the graph.
    \item We introduce a novel pipeline to generate the expressive spatio-temporal graphs from the video using a LMM.
    \item We also introduce key functions to perform multi-hop reasoning over the spatio-temporal graphs.
    \item Our comprehensive experiments and analysis demonstrate the effectiveness of our approach.
\end{itemize}

\section{Related Work}
\label{sec:related_work}
The field of video understanding has seen significant advancements, particularly with the integration of MLLMs. This section reviews key contributions and methodologies that have shaped the landscape of video comprehension.

\subsection{Large Multi-Modal Models}
Recent advances in Large Multi-Modal Models (LMMs) have demonstrated remarkable competencies in various tasks such as captioning and visual question answering. The LMMs like GPT-4V ~\cite{achiam2023gpt}, Gemini-1.5 ~\cite{team2024gemini}, and LLaMA 3.2 ~\cite{dubey2024llama} take a text prompt and a set of images as inputs and generate a rich text as an output which follows the input prompt as an instruction. 
Inspired by unprecedented capabilities of such LMMs, recent studies \cite{VideoAgent,longvu,malmm} have initiated explorations of extending LMMs for video understanding tasks. A primary challenge for the LMMs to understand the video contents is that it is impractical to process all the frames in the video since they typically convert a raw image into a sequence of tokens (visual tokens) using an image encoder ~\cite{vit} or vision-language models \cite{clip,blip2}. Due to their limited context length, they mostly can handle only few minutes of videos. This paper proposes a novel framework to address this limitation.


\subsection{Long-Term Memory for Video Understanding}
To address the limitation, various methods have been proposed for compressing long videos. Some techniques merge similar frames to create a long-term memory ~\cite{malmm,moviechat}, but they may miss crucial details and struggle with hallucinations. Later studies introduced query-aware memory ~\cite{moviechat_p,longvu}, which adaptively merges frames based on their similarity to an input query. For example, MovieChat+ ~\cite{moviechat_p} adjusts the compression ratio based on similarity, but this requires recompressing the video for each query, adding computational overhead. Other methods ~\cite{llovi,recap} divide the video into segments, generate textual descriptions for each, and store these as long-term memory. However, these descriptions might not capture essential details. Unlike these methods, we use a spatio-temporal graph to represent the video as long-term memory. This graph is generated once per video and retrieves key frames regardless of video length, allowing efficient processing of long videos.




\subsection{Graph as Structured Video Representation}
Representing videos as graphs has proven effective in various tasks, particularly in interaction detection \cite{video_sgg_1,video_sgg_2}, where nodes represent objects and edges capture their interactions.  This structured representation allows for capturing spatio-temporal relationships. Some studies \cite{video-of-thought,graph_video} have explored using graphs for video understanding, but their models may struggle with long videos and generalization due to training on specific datasets. We propose a novel pipeline to generate expressive spatio-temporal graph from a video with a LMM. Instead of feeding the graph into LLM, we run pre-designed reasoning functions over the graph to retrieve the key frames.

\begin{figure*}[!t]
    \centering
    \includegraphics[width=1.0\textwidth]{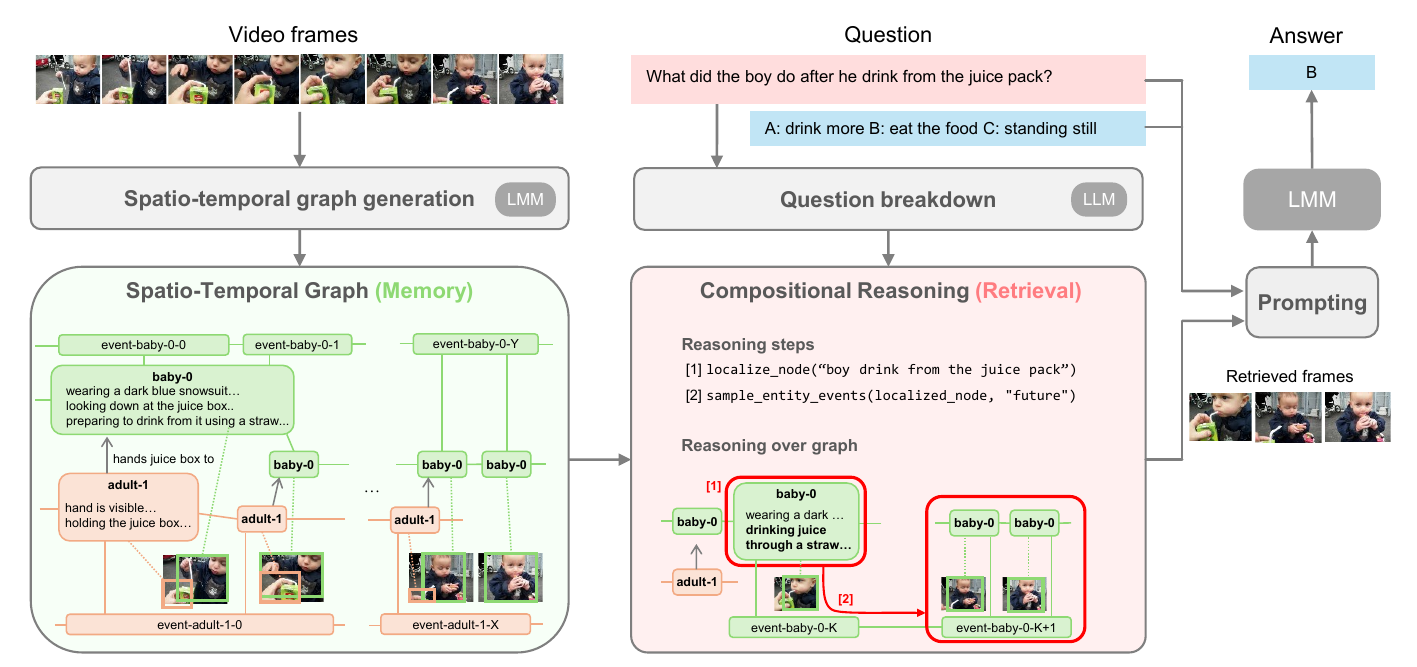}
    \caption{An entire pipeline of our RAVU consisting two core components: spatio-temporal graph generation (memory) and compositional reasoning (retrieval). These memory and retrieval mechanisms form the core of our proposed framework. Through the reasoning over the graph, the relevant frames can be identified and they are fed into the LMM to get a final answer for the given question.}
    \label{fig:pipeline}
\end{figure*}

\subsection{Multi-Step Reasoning for Video QA}
Some recent studies ~\cite{VideoAgent,video_rag} also have explored approaches to retrieve key frames relevant to the input query from a long video instead of compressing the video.
Agentic approaches like VideoAgent ~\cite{VideoAgent} iteratively retrieve relevant frames until sufficient information is gathered. However, these methods rely heavily on simple similarity between the query and individual frames, lacking the capability to track entities across frames. This limitation hinders their ability to handle complex queries that require multi-step reasoning to understand temporal relationships. To handle such multi-step reasoning, ~\cite{video-of-thought} has designed a specific reasoning steps. Unlike the existing works, we decompose the complex query into a sequence of reasoning steps and then execute each step with the reasoning function we designed for video understanding to retrieve the key frames.

\section{Method}

\begin{figure*}[t]
    \centering
    \includegraphics[width=1.0\textwidth]{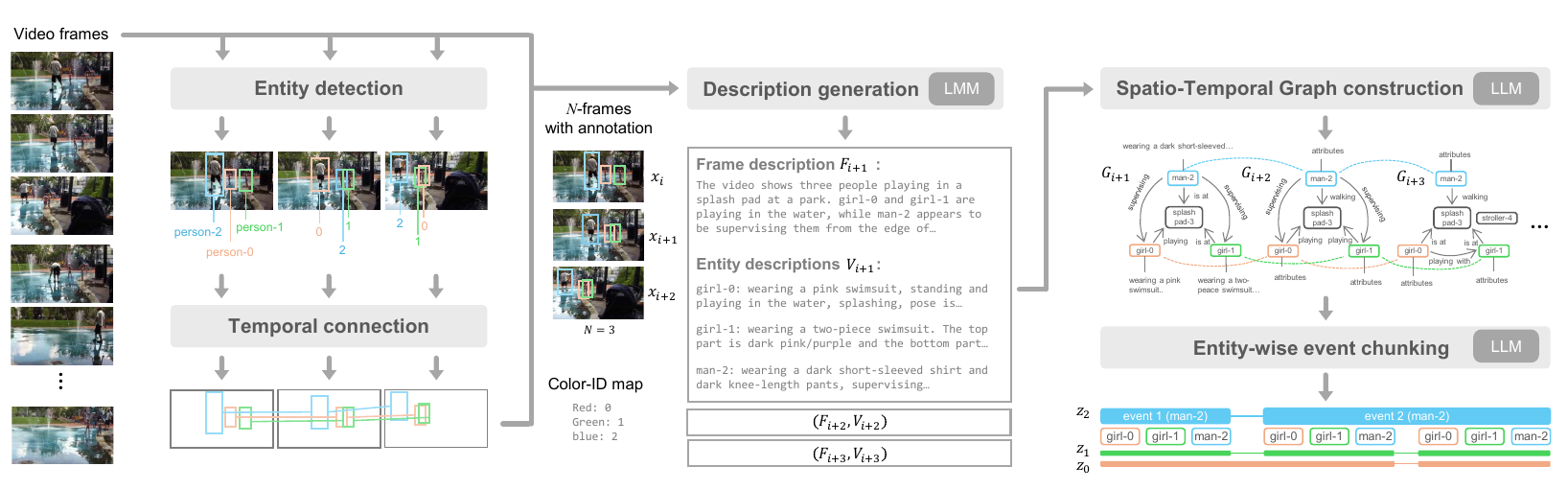}
    \caption{Spatio-temporal graph generation pipeline. Firstly, we detect entities in each frame and connect the detected entities across the frames, followed by LMM generates expressive descriptions for the frame and the entities with bounding box annotation. Then, we use LLM to construct the graph from them. The entity-wise events in the video are also generated to capture the segments of the video.}
    \label{fig:graph-gen}
\end{figure*}

In this section, we introduce our novel video understanding framework called \textbf{RAVU} (\textbf{R}etrieval \textbf{A}ugmented \textbf{V}ideo \textbf{U}nderstanding) and discuss our approach to VideoQA through compositional reasoning over graphs in detail. 
\subsection{Overview}

Figure \ref{fig:pipeline} illustrates an overview of our proposed framework, RAVU which consists of two key components: (1) a spatio-temporal graph generation module that constructs a structured representation of the video content (memory), and (2) a compositional reasoning module that operates over these graphs to localize relevant segments in response to user queries (retrieval). These memory and retrieval mechanisms constitute the foundation of RAVU.

\subsection{Spatio-Temporal Graph Generation}
We represent a video as a spatio-temporal graph, where each frame is modeled as a sub-graph comprising entity nodes and their relationships as edges. The entity nodes contain their visual attributes and spatial location in the frame. Nodes corresponding to the same entity across consecutive sub-graphs are connected to track the entities through time and capture their temporal dynamics.
While prior work has explored scene graph generation from images and videos ~\cite{sgg_survey,video_sgg_1,video_sgg_2}, these methods often suffer from limited vocabulary and poor generalization due to the constraints imposed by the small size of the training data. To address these limitations, we employ an LMM (Large Multi-modal Model) to generate frame-wise graphs and utilize object tracklets to establish temporal connections across frames. However, our preliminary experiment revealed two key challenges. First, directly prompting the LMM to identify entities, their attributes, and relationships within a frame often yields low-quality graphs, as the model tends to use the limited relation vocabulary. Second, establishing consistent correspondence between the entities across frames is not straightforward. This difficulty arises because visually similar entities might actually be distinct instances, leading the LMM to assign different IDs to the same entities across different frames. 

We propose a multi-step approach to enhance the robustness of the spatio-temporal graph generation, as shown in Fig.\ref{fig:graph-gen}. Our approach first generates expressive descriptions of video frames and entities in each frame, then converts these descriptions into a spatio-temporal graph. Specifically, we begin by detecting entities in each frame and tracking them based on bounding box matching to assign consistent IDs across the frames.
Next, we annotate the entities with distinct color-coded bounding boxes for every frame and prompt the LMM to generate the frame description, referring the entities by their IDs. We include a bounding-box color-to-ID map in the prompt, enabling the LMM to associate the bounding-box colors with the tracked identities. To generate rich and precise descriptions, we feed the LMM with a sequence of $N$ frames, consisting of a target frame and its neighboring frames. Finally, we prompt a LLM (Large Language Model) to convert the consistent frame descriptions and entities into a graph for each frame to construct a spatio-temporal graph.

More formally, we process each frame $x_i$ $(i=(0,...,N-1))$ annotated with the entities using the LMM $g_m$ with the instructions provided in the system prompt $s_{fd}$. This generates a tuple $(V_i, F_i)=g_m(s_{fd}, x_i)$, where, $V_i$ is the set of entities $\{n_i^{j}\}$ ($j\in\{0,...,M_{i}\}$) in the frame-$i$. Here, $M_i$ denotes the cardinality of the set $V_i$.
$F_i$ is the frame description generated by referring to the entities by their IDs.
Subsequently, for each frame $F_i$, LLM $g_l$ converts the consistent frame descriptions and entities into a graph $G_i(V_i, E_i)=g_l(V_i, F_i, s_g)$. Here, $E_i$ contains the edges representing the relationships between the entities from the frame description $F_i$. $s_g$ represents the system prompt with instructions to construct the graph for each frame. Note that each node contains rich attributes and its location obtained in the first step. Additionally, the prompt in this step utilizes only the entities and the frame description in textual format, excluding the video frame itself. %

To facilitate temporal reasoning, we further augment the graph by creating entity-wise events. This process involves chunking the spatio-temporal nodes for each entity into distinct events, thereby capturing significant behavioral and action changes. Specifically, for the entity-$j$, we generate the events $z^j$ as 
$z^j=g_l(\{p_i^j\}_{i=0}^{N-1}, s_e)$, where the system prompt $s_e$ contains instructions to segment the node into events, capturing the entity's actions and behaviors across the frames.



\subsection{Compositional Reasoning over Graph}
In this section, we present our methodology for localizing segments of a target video pertinent to answering a given question by reasoning over a spatio-temporal graph of the video. We employ dual-system models ~\cite{dual-system} in cognitive science to handle complex questions which require multi-step reasoning to identify the relevant segments. Our approach begins by breaking down the complex question into a sequence of reasoning steps. Then, we perform each reasoning step with a predefined function in sequence. Unlike existing works ~\cite{vipergpt,adacoder}, our reasoning functions operate on a spatio-temporal graph.

We first break down the complex questions using a LLM. To facilitate this process, we manually create examples of question analysis and breakdown for various types of questions, including temporal, descriptive, and causal, then use these examples in-context when analyzing new questions. The primary functions within our set of predefined actions include \emph{localize\_node} and \emph{sample\_entity\_events}. See our complete set of functions in Supplementary.


\begin{figure}[t]
    \centering
    \includegraphics[width=0.48\textwidth]{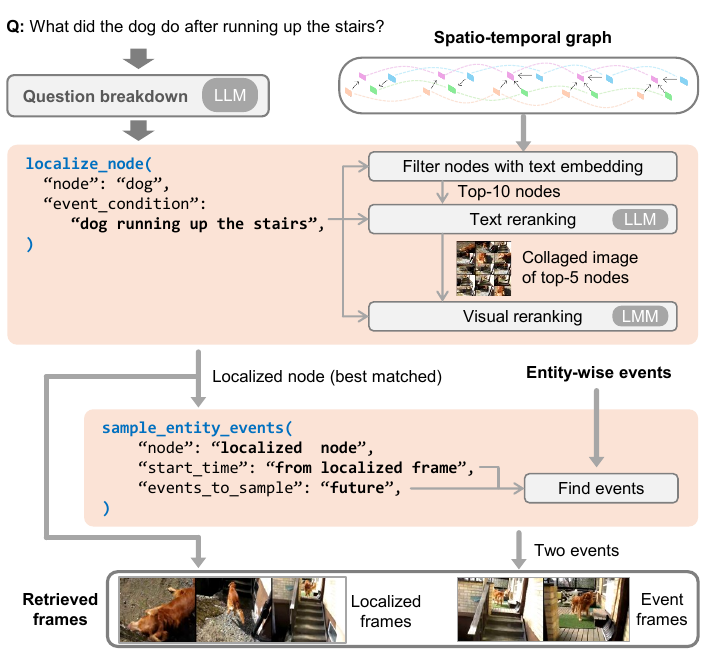}
    \caption{An example of our frame retrieval process. We execute reasoning steps sequentially and concatenate frames from each step.}
    \label{fig:retrieval}
\end{figure}

After breaking down the question into the sequence of reasoning steps, we perform them sequentially to retrieve the query-relevant key frames. We illustrate our retrieval process through an example in Figure \ref{fig:retrieval}. The question often requires multi-step reasoning to identify the relevant frames. Specifically, if the question is "What did the dog do after running up the stairs?", the spatio-temporal entity node “dog” with the attribute “dog running up the stairs” must first be identified. The function \emph{localize\_node} identifies this spatio-temporal node, finding the best match of the phrase $p_g$=\emph{“dog running up the stairs”} from the among the set of nodes $\{n_i^j\}$. 
We compute node embeddings by encoding concise textual description of the nodes with a text embedding model $g_e$. We first obtain the textual description $p_i^j$ of the node $n_i^j$ as, 
\begin{equation*}
    p_i^j=g_l(n_i^j, \{e_i\}, s_d),
\end{equation*}
where, $\{e_i\}$ is the set of all edges in frame-$i$ that include  entity-$j$ and $s_d$ is the system prompt with instructions to compose a sentence that encapsulates the entity node's attributes and its relations to other entity nodes. We then compute embeddings of each node as $v_i^j=g_e(p_i^j)$.
To localize the node, we first select the top-k spatio-temporal nodes whose embeddings exhibit the highest cosine similarity with the grounding phrase embeddings $v_g=g_e(p_g)$. Let $\eta=\{n_0, ..., n_{k-1}\}$ be the top-k filtered nodes and $P=\{p_0, ..., p_{k-1}\}$ be the corresponding node textual descriptions. Subsequently, we process the textual descriptions of the selected node to find the best matching node as $\hat{k}=g_l(P, p_g, s_r)$, where $s_r$ is the system prompt with instructions to select the phrase that best matches the grounding phrase and $n_{\hat{k}}$ is the best matched node. Let $n_{\hat{k}}$ correspond to the node of entity-$j$ in frame-$i$. Finally, the function \emph{sample\_entity\_events} is invoked to retrieve frames from events of entity-$j$ preceding the time index $t_e$.



\section{Experimental Settings}
\label{settings}
In this section, we provide a comprehensive description of the datasets utilized for evaluating our method. Subsequently, we elaborate on the various baselines and the implementation details.

\subsection{Evaluation Dataset}
We benchmark our method and compare it with various baselines and SOTA methods on two popular datasets, Next-QA ~\cite{nextqa} and EgoSchema ~\cite{egoschema}. 

\noindent \textbf{NExT-QA}: We follow ~\cite{VideoAgent} to focus on zero-shot evaluation on the validation set of the NExT-QA. It contains 570 videos and 5,000 multiple choice questions. NExT-QA provides 8 types of questions, including 2 types of causal questions, 3 types of temporal question, and 4 types of descriptive questions. Especially, this paper aims to address the temporal next/previous questions which are particularly difficult and require multi-hop reasoning to infer the past or future. These types of questions compose 29\% of the dataset. The videos in this dataset average 45 and maximum 180 seconds in length.

\noindent \textbf{EgoSchema}: EgoSchema is a benchmark for zero shot comprehension of the long-form videos, containing 5,000 multiple-choice questions based on 5,000 egocentric videos. These videos capture a first-person perspective of individuals participating in various activities. Due to the focus on zero-shot evaluation, this dataset only comprises of the test set. Each video in this dataset is 3 minutes long. We follow ~\cite{VideoAgent} and evaluate on a subset of this dataset containing 500 questions corresponding to 500 videos having publicly accessible labels. 

We employ accuracy as our evaluation metric since the datasets features multiple-choice questions.

\subsection{Baselines and Other Methods}
We compare RAVU with other state-of-the-art methods including both supervised and zero-shot methods, such as LongVU \cite{longvu}, LLoVi ~\cite{llovi}, and VideoAgent ~\cite{VideoAgent}. However, it is not easy to compare our method with other existing zero-shot methods since they rely on different proprietary models (e.g., GPT-3.5 and GPT-4). Therefore, for a fair comparison, we conduct following exhaustive baseline experiments with a specific LMM as a fixed reasoning model across all the experiments.

\begin{itemize}
    \item \textbf{BlindQA}: We just feed the multiple choice question (MCQ) and do not feed the video frames to the LMM.
    \item \textbf{All frames}: We feed all the frames (at 1fps) to the LMM to answer the question. 
    \item \textbf{Image-based frame retrieval}: Here we employ a CLIP model to retrieve the top-$5$ frames most relevant to the query based on text-image similarity. These selected frames were passed to the LMM model for answering the question.  

    \item \textbf{Text-based frame retrieval}: We use the frame descriptions as the retrieval key instead of the frame and compute the similarity to the query in the embedding space to find the top 5 relevant frames. These query relevant frames are then passed to the LMM for the question answering task. 


\end{itemize}

\subsection{Implementation Details}

We use \textit{gemini-1.5-flash-002} for detecting entities, generating the frame descriptions, constructing a graph, question breakdown, and inferring the final answer for the questions for all the experiments. This is a Gemini Flash model, and it is cheaper and faster than the Gemini Pro model, and thus can be used more often in the realistic scenarios. For entity tracking, we use SAM2 ~\cite{sam2} as a tracker for EgoSchema and annotation in VidOR dataset ~\cite{vidor} for NExT-QA. We also use SAM2 for NExT-QA for comparison.
For retrieval, we use a Sentence Transformer ~\cite{sentencebert} of \textit{all-mpnet-base-v2} model. Video frames are uniformly sampled from the videos at 1 fps. For baseline experiments, we use \textit{EVA02-CLIP-L-14} for text-image retrieval and Gemini \textit{text-embedding-004} model for text retrieval. We set all the safety filters to the lowest option, while 33 videos were blocked. 

Recent approaches ~\cite{llovi,VideoAgent} uses question options to retrieve the relevant frames or texts since these can serve as key words that directly correspond to the related images or the text, while this setting may not be a realistic scenario and thus we do not use them for retrieval.

\begin{table}[t]
\setlength{\tabcolsep}{2.0pt} 
\footnotesize 
\small
\centering
\begin{tabular}{lcccc}
\toprule
Models & $Acc_C$ & $Acc_T$ & $Acc_D$ & $Acc$ \\
\midrule
Human & 87.61 & 88.56 & 90.4 & 88.38 \\
\midrule
\multicolumn{5}{c}{\textit{Supervised}}\\
HiTeA ~\cite{HiTeA} & 62.4 & 58.3 & 75.6 & 63.1 \\
VFC ~\cite{momeni2023verbsactionimprovingverb} & 49.6 & 51.5 & 63.2 & 52.3 \\
Vamos ~\cite{vamos} & \textbf{77.2} & \textbf{75.3} & 81.7 & \textbf{77.3} \\
SeViLA ~\cite{yu2023selfchainedimagelanguagemodelvideo} & 73.8 & 67.0 & 81.8 & 73.8 \\
MotionEpic ~\cite{fei2024videoofthought} & 75.8 & 74.6 & \textbf{83.3} & 76.0 \\
VLAP ~\cite{DBLP:journals/corr/abs-2312-08367} & 74.9 & 72.3 & 82.1 & 75.5 \\
ViLA ~\cite{ViLA} & 75.3 & 71.8 & 82.1 & 75.6 \\
\midrule
\multicolumn{5}{c}{\textit{Zero-shot}}\\

AssistGPT ~\cite{gao2023assistgptgeneralmultimodalassistant} & 60.0 & 51.4 & 67.3 & 58.4 \\
SeViLA ~\cite{yu2023selfchainedimagelanguagemodelvideo} & 61.3 & 61.5 & 75.6 & 63.6 \\
ViperGPT ~\cite{vipergpt} & - & - & - & 60.0 \\
LLoVi ~\cite{llovi} & 67.1 & 60.1 & 76.5 & 66.3 \\
VideoAgent ~\cite{VideoAgent} & 72.7 & 64.5 & \textbf{81.1} & 71.3 \\
MotionEpic ~\cite{fei2024videoofthought} & - & - & - & 66.5 \\
\rowcolor{blue!15}
RAVU (non-blocked content) & \textbf{76.67} & \textbf{68.91} & 76.11 & \textbf{74.09} \\
\rowcolor{blue!15}
RAVU (overall) & 74.40 & 66.56 & 74.64 & 71.93 \\
\bottomrule
\end{tabular}
\caption{Results on NExT-QA for Supervised and Zero-shot state-of-the-art methods. $Acc_C$, $Acc_T$, $Acc_D$ and $Acc$  represent accuracy on causal, temporal, descriptive subsets and overall accuracy, respectively. We bold the best results.}
\label{tab:nextqa_part1}
\end{table}

\section{Results and Analysis}

\subsection{Comparison with State-of-the-arts}

\noindent \textbf{NExT-QA}: 
In Table \ref{tab:nextqa_part1}, we present the performance of our proposed RAVU model alongside other state-of-the-art (SOTA) supervised and zero-shot video understanding methods on the NeXT-QA dataset. It is important to note that our approach incorporates the proprietary Gemini model as a foundational component. Due to the non-configurable safety protocols embedded within Gemini, certain questions, frames, or videos identified as unsafe content are consequently blocked. Specifically, for the NeXT-QA dataset, our evaluation was conducted on 4,856 out of 5,000 questions, with the remaining questions being blocked by Gemini. To ensure a fair comparison with other SOTA methods, we report the accuracy of RAVU on both the non-blocked questions and on all questions, under the assumption that the blocked questions are incorrect. Notably, despite the limited number of frames (an average of 5 frames per video), RAVU demonstrates competitive performance relative to other methods. 

\noindent \textbf{EgoSchema}: EgoSchema comprises global behavioral questions that necessitate reasoning over entire videos. For such questions, we employ a hierarchical retrieval approach. Our method utilizes a spatio-temporal knowledge graph, which includes entity node descriptions and event segmentations for each entity. Initially, we retrieve the most relevant entity node descriptions from each event based on the similarity between the embeddings of the question and the frame descriptions within that event. We then prompt the LMM with these retrieved descriptions to select the top 10 descriptions that best match the query. The frames corresponding to these top 10 descriptions are subsequently fed to the LMM along with the query to generate the answer. This event-based approach ensures diversity and relevance in the sampled frames. In Table \ref{tab:egoschema_part1}, we present the performance of RAVU and other state-of-the-art (SOTA) supervised and zero-shot video understanding methodologies on the EgoSchema dataset. For the EgoSchema dataset, six questions were blocked by the Gemini model, resulting in an evaluation on 494 questions for the non-blocked setting. We observe that RAVU demonstrates competitive performance with just 10 retrieved frames.

\begin{table}[t]
\small
\centering
\begin{tabular}{lc}
\toprule
Methods & $Acc$\\
\midrule
\multicolumn{2}{c}{\textit{Supervised}}\\
LongViViT ~\cite{papalampidi2024simple} & 56.8\\ 
MC-ViT-L ~\cite{MC-ViT-L} & 62.6\\ 
\midrule
\multicolumn{2}{c}{\textit{Zero-shot}}\\
SeViLA ~\cite{yu2023selfchainedimagelanguagemodelvideo} & 25.7\\ 
LLoVi ~\cite{llovi} &  52.2\\ 
VideoAgent ~\cite{VideoAgent}  & 60.2\\ 
\rowcolor{blue!15}
RAVU (non-blocked content) & \textbf{67.41} \\
\rowcolor{blue!15}
RAVU (overall) &  66.60 \\
\bottomrule
\end{tabular}
\caption{Results on EgoSchema 500 video subset 
 as compared to state-of-the-art methods.We bold the best results.}
\label{tab:egoschema_part1}
\end{table}

\begin{table*}[!t]
\setlength{\tabcolsep}{5.0pt}
\begin{center}

\scalebox{1.0}{
\begin{tabular}{lccccccccccccc}
\toprule
\multirow{2}{*}{Methods} & \multicolumn{3}{c}{$Acc_C$} & \multicolumn{4}{c}{$Acc_T$} & \multicolumn{4}{c}{$Acc_D$} & \multirow{2}{*}{$Acc$} & Cost\cr
\cmidrule(lr){2-4} \cmidrule(lr){5-8} \cmidrule(lr){9-12}
& CW & CH & $All_{C}$ & TP & TN & TC & $All_{T}$ & DC & DL & DO & $All_{D}$ & & \(\left(10^3\right)\)\cr
\midrule
BlindQA (only MCQs) & 35.3 & 38.1 & 36.0 & 13 & 17.3 & 21.6 & 18.9 & 1.1 & 8.1 & 19.0 & 10.6 & 26.6 & 0.2\cr
All frames (1 fps) & 80.1 & 81.0 & 80.3 & 69.6 & 71.6 & 79.6 & 75.1 & 64.9 & 87.7 & 89.1 & 82.5 & 79.0 & 11.8\cr
\midrule
Clip-based retrieval (k=5) & 74.2 & 76.2 & 74.7 & 70.2 & 59.2  & 72.5 & 65.0 & 51.1 & \textbf{90.9} & 82.0 & \textbf{77.7} & 72.3 & 1.4\cr
Text-based retrieval (k=5) & 74.2 & 76.7 & 74.4 & 70.2 & 59.0 & 73.7 & 65.4 & \textbf{52.0} & 89.3 & 83.5 & 76.9 & 72.6 & 1.4\cr
\rowcolor{blue!15}
RAVU (ours) & \textbf{76.7} & \textbf{77.1} & \textbf{76.8} & \textbf{78.7} & \textbf{64.0} & \textbf{75.5} & \textbf{69.2} & 42.8 & 89.8 & \textbf{85.4} & 76.5 & \textbf{74.6} & 5.9\cr
\bottomrule
\end{tabular}
}

\caption{Zero-shot performance comparison with baselines on \textbf{NExT-QA} with Gemini 1.5 Flash as a reasoning LMM. $Acc_C$, $Acc_T$, and $Acc_D$ are accuracy on causal, temporal, and descriptive subsets, respectively. CW/CH: causal-why/how, TP/TN/TC: temporal previous/next/current, DC/DL/DO: descriptive count/location/others.We bold the best results. \textbf{Cost} denotes the input token count per question.}
\label{tab:nextqa_part2}

\end{center}
\end{table*}

\begin{table}[t]
\small
\centering
\begin{tabular}{lc}
\toprule
Methods & $Acc$\\
\midrule
All Frames (1 fps) & 70.67\\ 
\midrule
CLIP-based retrieval &  63.88\\ 
\rowcolor{blue!15}
RAVU (ours) & \textbf{67.76} \\
\bottomrule
\end{tabular}
\caption{Zero-shot performance comparison with baselines on EgoSchema 500 video subset. We bold the best results.}
\label{tab:egoschema_part2}
\end{table}

\subsection{Comparison with baselines}

\noindent \textbf{NExT-QA}: We present the performance of the baseline methods and our proposed approach in Table \ref{tab:nextqa_part2}. For a fair evaluation, all methods in this table were assessed on 4,596 questions that were not blocked by the Gemini model for any of the methods. We note that the proposed RAVU approach demonstrates superior performance compared to other retrieval-based baselines, while utilizing a similar number of frames to answer the questions. Notably, we observe significant performance improvements in the temporal category of questions. This is anticipated, as temporal questions often involve queries about the state of entities following or preceding the event in question, which are challenging to address with similarity based retrieval methods. Additionally, we report the cost for each method in terms of the average number of tokens per question for the NExT-QA dataset. Our proposed approach incurs a higher cost than other methods due to the query breakdown process and use of LMM in the retrieval process. Query breakdown incurs more than half of the cost with 3465 tokens per question due to in-context query breakdown illustrations. However, this cost can be significantly reduced through finetuning the LMM for query breakdown which will remove the need of in-context examples. Further, the cost of reasoning in retrieval can also be reduced through system prompt compression techniques \cite{gist}.

\noindent \textbf{EgoSchema}: We present the performance for the EgoSchema dataset in Table \ref{tab:egoschema_part2}. For the frame retrieval baseline, we employed the CLIP model to retrieve the top 10 frames for each video. To ensure a fair evaluation, all methods in Table \ref{tab:egoschema_part2} were assessed on 490 questions from an equal number of videos, which were not blocked by the Gemini model for any of the methods. We note that RAVU exhibits superior performance compared to the frame retrieval-based baseline. This underscores the efficacy of our approach in retrieving frames for global behavioral question types when compared to the similarity based retrieval approach.


\subsection{Localization Analysis}
In this section, we evaluate the accuracy of our frame localization methodology within the $localize\_node$ function. Our approach, particularly for causal and temporal questions, involves initially localizing the entity and event referenced in the question. Subsequently, we sample from future, past, or neighboring events based on the requirements of the question. Therefore, assessing the localization performance is crucial, as subsequent processes depend on accurate localization. To this end, we manually annotated 381 questions from 49 randomly selected videos from the NExT-QA dataset. Specifically, for each question, we identified the frames containing the event mentioned in the question, excluding frames depicting events occurring before or after the specified event.

To evaluate our method on this data, we compared the frame indices predicted by the $localize\_node$ function to the ground truth frame indices. If the predicted frame indices fall within the ground truth, we consider the prediction correct, otherwise, it is deemed incorrect. 
We compare the localization performance of CLIP Embeddings, text embeddings of entity node descriptions and our proposed reranking approach for localization. We report the localization results in Table \ref{tab:localization}. We note that our proposed approach results in significant localization performance gains when compared to other approaches.  



\begin{table}[!t]
\begin{center}
\scalebox{1.0}{
\begin{tabular}{lcc}
\toprule
Methods & $Acc_C$ &$Acc_T$ \cr
\midrule
Proposed Reranking& \textbf{70.57}  & \textbf{58.15} \cr
Text Embedding & 58.69  & 44.05 \cr
CLIP Embedding & 60.87  & 44.47 \cr
\bottomrule
\end{tabular}
}




\caption{Frame localization accuracy with different ranking algorithms on a subset of NExT-QA on causal and temporal questions.}
\label{tab:localization}
\end{center}
\end{table}

\subsection{Impact of Generated Graphs}
To evaluate our graph generation methodology, we measure QA accuracy on 424 questions from our subset of 49 videos our using three distinct graph variants: (1) a graph generated by our approach utilizing ground-truth tracklets from the VidOR dataset, (2) a graph generated by our approach using tracklets predicted by SAM2, and (3) a graph constructed from human-annotated scene graphs in the VidOR dataset. We report the results in Table \ref{tab:compare_graph}. The highest accuracy is achieved using the spatio-temporal graphs from the VidOR dataset. This suggests that entity tracking performance may be a critical bottleneck. While our graph demonstrates greater expressiveness compared to the VidOR graph, which has a limited vocabulary, its performance is slightly inferior. This can be attributed to the presence of hallucinations in our graph, which is generated by the LLM, unlike the human-annotated graphs from VidOR.

\begin{table}[t]
\setlength{\tabcolsep}{4.0pt}
\small
\centering
\begin{tabular}{lcccc}
\toprule
Methods & Acc@C & Acc@T & Acc@D & Acc@All \\
\midrule
w/ GT tracklets & 76.16 & 74.19 & 73.58 & 75.29 \\
w/ SAM2 tracklets & 77.51 & 70.86 & 73.58 & 74.58 \\
w/ VidOR annotation & 78.24 & 75.16 & 74.54 & 76.65 \\
\bottomrule
\end{tabular}
\caption{Zero-shot performance ablations with different graphs on a subset of NExT-QA.}
\label{tab:compare_graph}
\end{table}

\section{Conclusion}

We introduced RAVU, a novel retrieval-augmented video understanding framework that constructs spatio-temporal graphs for long-term memory and compositional reasoning. By leveraging these graphs for frame retrieval, RAVU excels in addressing complex temporal, causal, and global reasoning tasks. Evaluations on NExT-QA and EgoSchema show superior performance in answering multi-hop and object-tracking queries with minimal frame retrieval, highlighting its effectiveness in video understanding.




\section*{Ethical Statement}

This study does not involve any ethical concerns. All datasets used in this work, including NExTQA and EgoSchema, are publicly available and and no sensitive or personal information is used.




\clearpage
\newpage

\bibliographystyle{named}
\bibliography{ijcai25}

\clearpage
\newpage
\appendix
\section{Reasoning Functions}
Table \ref{tab:func_list} provides a list of reasoning functions we design to handle multi-hop reasoning over a spatio-temporal graph.

\begin{table}[t]
\small
\centering
\begin{tabular}{lcc}
\toprule
Functions & Arguments & Descriptions\\
\midrule
$localize\_node$ & query & retrieves the most relevant node and corresponding frame\\
$sample\_entity\_events$ & node, sample\_start\_time, events\_to\_sample & sample frames from relevant entity events\\
$extract\_temporal\_part$ & target\_part & extracts relevant video segment (beginning, middle or end)  \\
$count\_nodes$ & node, event\_condition & called for counting questions\\
$get\_global\_context$ & - & samples frames uniformly\\
$analyze\_events$ & query & LMM analyzes events for temporal reasoning \\
$identify\_node$ & query  & uses LMM to identify entity node based on given query\\
\bottomrule
\end{tabular}
\caption{A list of our reasoning functions.}
\label{tab:func_list}
\end{table}

\end{document}